# Active Learning with Distributional Estimates

Jens Röder[1], Boaz Nadler[2], Kevin Kunzmann[3], and Fred A. Hamprecht[3]

[1]CR/AEM, Robert Bosch GmbH, Hildesheim, Germany
[2]Department of Computer Science and Applied Math, Weizmann Institute of Science, Rehovot, Israel
[3]Heidelberg Collaboratory for Image Processing (HCI), University of Heidelberg, Heidelberg, Germany

## Abstract

Active Learning (AL) is increasingly important in a broad range of applications. Two main AL principles to obtain accurate classification with few labeled data are *refinement* of the current decision boundary and *exploration* of poorly sampled regions. In this paper we derive a novel AL scheme that balances these two principles in a natural way. In contrast to many AL strategies, which are based on an estimated class conditional probability $\hat{p}(y|x)$, a key component of our approach is to view this quantity as a random variable, hence explicitly considering the *uncertainty* in its estimated value. Our main contribution is a novel mathematical framework for uncertainty-based AL, and a corresponding AL scheme, where the uncertainty in $\hat{p}(y|x)$ is modeled by a second-order distribution. On the practical side, we show how to approximate such second-order distributions for kernel density classification. Finally, we find that over a large number of UCI, USPS and Caltech-4 datasets, our AL scheme achieves significantly better learning curves than popular AL methods such as uncertainty sampling and error reduction sampling, when all use the same kernel density classifier.

## 1 INTRODUCTION

In many applications, including computer vision and natural language processing, unlabeled data abounds while procuring labels for training is costly. Pool-based active learning (AL) schemes judiciously select those among the unlabeled points that are deemed most informative, and thought to help achieve a steeper learning curve. The prospect of reduced labeling effort has spurred intense efforts to improve AL. On the theoretical side, several works considered the sample complexity and potential benefits of AL, see (Beygelzimer et al., 2009; Balcan et al., 2010; Hanneke, 2011). On the practical side, various works suggested concrete AL schemes, recently e.g. (Huang et al., 2010; Siddiquie and Gupta, 2010), with large gains over random labeling in various applications, see (Settles, 2010) for a comprehensive review.

In this paper we focus on pool based AL. We first review a potential weakness common to many popular AL methods, and then derive a new pool-based AL scheme. In a pool-based setting, one typically starts with a small (possibly empty) set of labeled samples $\mathcal{L} = \{x_i, y_i\}_{i=1}^{\ell}$, and a large pool of unlabeled samples $\mathcal{U} = \{x_j\}_{j=\ell+1}^{n}$. Most pool-based AL schemes rely on a classifier – or more precisely, a regressor – that outputs not only a predicted class label $\hat{y}$ at a new sample $x$, but also an estimate $\hat{p}(y|x)$ of the conditional class probabilities $\Pr[Y = y|X = x]$ for all classes $y$. Then, sequential one-step looka-head AL schemes compute a Training Utility Value (TUV) for any unlabeled sample, and query the label of the sample with largest TUV.

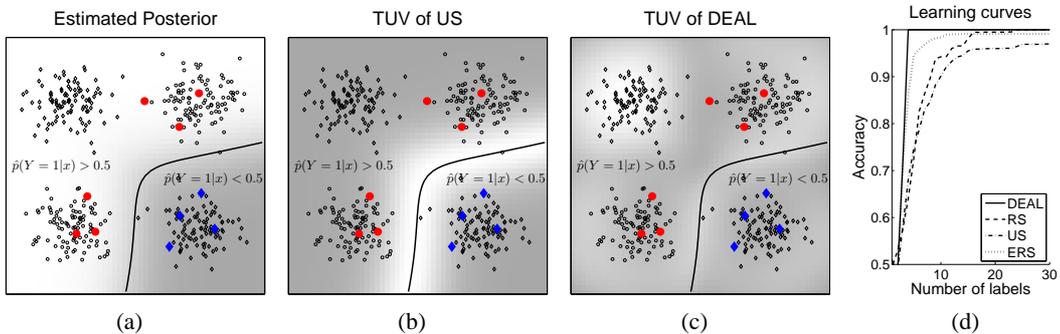

Figure 1: Active learning on the XOR problem. Small black symbols are unlabeled data, large colored symbols are the labeled ones. (a) Class-conditional probability and decision boundary estimated by kernel density classification, with labeled data in only 3 out of 4 quadrants. (b), (c) Training utility values (TUV) of uncertainty sampling and the proposed DEAL, lighter color representing a higher TUV. Uncertainty sampling prematurely concentrates on local refinement of the current decision boundary. DEAL keeps exploring before reverting to refinement. (d) Resulting learning curves.

One popular and successful AL strategy is uncertainty sampling (US) which iteratively selects the sample whose current class prediction is least confident[1] (Baum, 1991; Hwang et al., 1991; Seung et al., 1992; Lewis and Gale, 1994).

Another common strategy is to query that sample whose inclusion in the training set may contribute most towards a "confident" classification. Here, confidence is measured by the entropy of the class conditional probabilities, or the expected estimated risk (MacKay, 1992; Roy and McCallum, 2001; Zhu et al., 2003): the more assertive the resulting classifier, the better, according to these algorithms. A variant in the regression context has been proposed in (Boutilier et al., 2003).

Yet another popular AL strategy is to select samples that minimize the uncertainty in the estimated parameters of a classifier, e.g. by maximally reducing the version space of a SVM (Tong and Koller, 2002) or by minimizing the variance of the parameter estimates in multinomial logistic regression (Schein and Ungar, 2007).

A common theme to all these AL schemes is their use of point estimates $\hat{p}(y|x)$, possibly combined with density estimates $\hat{p}(x)$, but without consideration of the inherent random uncertainties in these quantities. By definition, $\hat{p}(y|x)$ is estimated from the finite, and often small, currently labeled set $\mathcal{L}$.[2] Hence, at any $x$, $\hat{p}(y|x)$ is a random variable, which may have small bias and variance in some regions, but high uncertainty in others.

In this paper, we propose to capitalize on this seeming flaw, and to put the unavoidable uncertainty in the estimates $\hat{p}(y|x)$ at the very heart of a novel AL scheme: Distributional Estimate Active Learning (DEAL). First, in Section 2 we propose to quantify the uncertainty in the estimate $\hat{p}(y|x)$ via a *second-order distribution*, see Eq. (1). Next, in Section 3 we show how such a second-order distribution can be approximated for kernel density classification; and in Section 4 we show how such distributions can be used, in a principled mathematical framework, for uncertainty-based AL. In Section 5 we show empirically that with a baseline implementa-tion using kernel density classification, DEAL performs significantly better than two highly popular AL schemes and random sampling in a thorough benchmark on more than 40 classification problems from the UCI (Frank and Asuncion, 2010) and USPS (LeCun et al., 1990) databases, and on an image classification task using the Caltech-4 dataset.

---

[1] Note that this can be defined in many possible ways, in particular in multi-class settings.

[2] And, in the case of semi-supervised active learning, also from the unlabeled set $\mathcal{U}$.

Our approach is somewhat related to the minimization of uncertainty in Gaussian process regression for space-filling experimental design (Sacks et al., 1989). In the machine learning community, several works devised efficient approximations for the intractable posterior distribution in Gaussian process classification models (Nickisch and Rasmussen, 2008). In particular, these distributions were used to compute Bayesian predictive distributions (Snelson and Ghahramani, 2005), which for classification are *point estimates* of the class conditional probabilities. Gaussian processes were also used for active learning, though there the authors suggested to label those samples whose normalized margin is smallest (Kapoor et al., 2007). Thus, even though second-order distributions were derived for logistic regression and Gaussian processes, to the best of our knowledge, these have not been used in AL for classification. In this paper we thus emphasize the importance, use, and potential benefit of second-order distributions in AL classification problems. As discussed in Section 6, second-order distributions may see potential use beyond AL.

## 2 CLASSIFIER UNCERTAINTY AND ACTIVE LEARNING

In statistical pattern recognition, agreement prevails that a classifier should not be forced to make a prediction unless reasonably confident about it. This principle is formalized by introducing an auxiliary "doubt" class that the classifier can always vote for, at a fixed cost (Ripley, 2008). In the generic case of a symmetric loss function, minimizing the expected risk leads to an algorithm that, given a sample $x$, votes for the class with highest conditional probability, $\hat{y} = \arg\max_y \hat{p}(y|x)$, provided that the expected loss of this decision is smaller than the fixed cost of the "doubt" class.

The "doubt" class captures the uncertainty of a prediction $\hat{y}$ at locations $x$ where no class is clearly dominant. Even if the class conditional probabilities are perfectly known, this type of "first-order" uncertainty is still present wherever two classes overlap in feature space. As a direct consequence to AL, if the current labeled set makes it quite clear that two classes are equally probable at some region in feature space, it is futile to attempt reducing this first-order uncertainty by requesting more labels there!

In practice, $p(y|x)$ is unknown, and thus estimated from a finite training set. This induces a *second* kind of uncertainty: not only how confident are we in the predicted label $\hat{y}$, but also how accurate is our point estimate $\hat{p}(y|x)$. An inaccurate point estimate may result in a misleading classifier that errs and votes for the wrong class, with a class conditional probability margin that is deceptively large. Asking for the label of additional training samples in such regions can result in a decisive change in the current decision boundary. Hence, samples with highly uncertain class conditional probability estimates should be prime candidates of a good AL criterion. A point in case is the classical XOR problem, illustrated in Fig. 1. Starting with 10 labeled data in only 3 out of 4 quadrants (an event whose probability is ∼20% with 10 randomly selected labeled data), nearest-neighbor type classifiers give an erroneous prediction at the remaining quadrant, with a deceptively large margin. Consequently, AL schemes based on $\hat{p}(y|x)$ do not sample points in the remaining quadrant. This overconfidence of AL schemes was also noted by (Baram et al., 2004), who suggest to label at random once in a while.

Motivated by the above insights, in this paper we derive an AL scheme that incorporates this randomness in $\hat{p}(y|x)$ in a natural way. The key ingredient in our scheme is a *second-order distribution*

$$G_x(q) = \Pr[\hat{p}(y=1|x) \leq q] \qquad (1)$$

which measures our uncertainty in the point estimate $\hat{p}(y|x)$. Before deriving the DEAL scheme, we first show how such a second-order distribution can be estimated for the kernel density classifier.

# 3  SECOND-ORDER DISTRIBUTIONS FOR THE KERNEL DENSITY CLASSIFIER

Kernel density classification is a prototypical non-parametric generative classifier. While with limited training data this classifier will likely have a lower accuracy compared to modern discriminative classifiers, we choose it since it is: ($i$) conceptually simple and easy to implement, ($ii$) usable in all the active learning criteria that we wish to benchmark and ($iii$) representative of an entire class of more advanced methods. While beyond the scope of this paper, second-order distributions can also be derived for discriminative classifiers, and then used in our AL scheme.

For simplicity, in the rest of this paper we focus on the binary classification problem, with class labels $y \in \{-1, +1\}$. To derive second-order distributions for the unknown class probabilities $p(1|x) = \Pr[Y=1|X=x]$, we use Bayes rule

$$p(1|x) = \frac{p(x|1)\pi_1}{p(x|-1)\pi_{-1} + p(x|1)\pi_1} \quad (2)$$

with $\pi_y$ the prior probability for class $y$. In kernel density classification, the unknown class densities $p(x|1)$ and $p(x|-1)$ are replaced by their Parzen window estimates. To derive second-order distributions, we thus need to approximate the distribution of these point estimates.

Let $\mathcal{K}$ be a normalized ($\int \mathcal{K}(u)du = 1$) isotropic kernel. Then the kernel density estimate

$$\hat{p}(x|Y=y) = \frac{1}{n_y} \sum_{x_i : y_i = y} \mathcal{K}(x - x_i) \quad (3)$$

is a random variable. With only a single observation from class $y$ ($n_y = 1$), the exact distribution of this random variable is given by

$$\Pr[\hat{p}(x|y) \leq z] = \int \mathbf{1}\left(\mathcal{K}(u-x) \leq z\right) p(u|y) du \quad (4)$$

This distribution depends on the location of the query $x$, on the kernel function $\mathcal{K}$ and on the unknown density $p(x|Y=y)$. Qualitatively, for non-negative and monotonically decaying kernels $\mathcal{K}$, the resulting density must be zero for $z < 0$ and for $z > \mathcal{K}(0)$. For $n_y$ i.i.d. observations from class $y$, the distribution of the class density kernel estimate is given by a $n_y$-fold convolution of the probability density that corresponds to Eq. (4).

Since the exact density $p(x|Y=y)$ is unknown (otherwise no learning would be necessary), we resort to an approximation of the distribution of $\hat{p}(x|y)$. Key requirements are that the approximate distribution be continuous, infinitely divisible, have no mass at $z < 0$ and its derivative should decay to zero as $z \to \infty$. A good candidate meeting these criteria is the Gamma distribution which, with its shape and location parameters $k$ and $\theta$, is also sufficiently rich to faithfully model a variety of situations that arise in practice. A standard estimate of mean and variance of the kernel density estimate (Härdle et al., 2004, chap. 3) allows to apply the method of moments and obtain the shape parameter $k_y = n_y \hat{p}(x|Y=y)/C_2$ and location parameter $\theta_y = C_2/n_y$, where $C_2 = \int \mathcal{K}^2(x)dx$.

When the class prior $\pi_y$ is estimated by the ratio $n_y/n$, we obtain the following approximate distribution for the random variable $\hat{p}(x|y)$

$$\hat{p}(x|y)\hat{\pi}_y \sim \Gamma\left(\delta + \frac{n_y \hat{p}(x|y)}{C_2}, \frac{C_2}{n_y} \cdot \frac{n_y}{n}\right). \quad (5)$$

Here, $\delta$ is a small positive constant added both to regularize Eq. (2) in low-density regions, and to guarantee that the shape parameter of the Gamma distribution is strictly positive, even when no labeled observations are available yet for class $y$.

With $\theta := C_2/n$, inserting Eq. (5) into Eq. (2) gives

$$\hat{p}(Y=1|x) \sim \frac{\Gamma(\delta + k_1, \theta)}{\Gamma(\delta + k_{-1}, \theta) + \Gamma(\delta + k_1, \theta)}$$
$$= Beta(\delta + k_1, \delta + k_{-1}) = Beta(\alpha, \beta) \quad (6)$$

In particular, for a $d$-dimensional isotropic Gaussian kernel with bandwidth $h$, we obtain

$$k_y = 2^{d/2} \sum_{x_i : y_i = y} \exp\left(-\|x - x_i\|^2 / 2h^2\right). \quad (7)$$

# 4 DISTRIBUTIONAL ESTIMATE ACTIVE LEARNING (DEAL)

Our novel AL scheme requires a method (for instance the one described in the previous section, or a Gaussian process classifier) that outputs a second-order distribution $G_x(q)$, Eq. (1).

Our point of departure in deriving our AL scheme, is the following key observation (Friedman, 1997): The performance of a classifier, as measured by its misclassification error, depends only on the location of its decision boundary, and not on the precise estimates of the conditional class probabilities. In particular, inaccurate point estimates $\hat{p}(y|x)$ may still yield the optimal Bayes classifier as long as they result in the same decision boundary.

A second-order distribution can thus help assess the uncertainty in the currently estimated decision boundary. In more detail, given a second-order distribution with density $g_x(q) = \frac{d}{dq} G_x(q)$, we can extract a point estimate for the posterior probability

$$\hat{p}(1|x) = \int_0^1 q \, g_x(q) dq \qquad (8)$$

and a corresponding classifier, which for a symmetric loss function is simply

$$\hat{f}(\hat{p}(1|x)) = \text{sgn}(\hat{p}(1|x) - 1/2). \qquad (9)$$

The goal of classification is to build a classifier $\hat{f}$ that minimizes the overall risk

$$R = \int R[x, \hat{f}] p(x) dx \qquad (10)$$

where the local risk at $x$ is

$$R[x, \hat{f}] = \mathbb{E}_Y[L(y, \hat{f})] = \sum_{y=\pm 1} L(y, \hat{f}) p(Y=y|x). \qquad (11)$$

In general, the exact local risk at $x$ is unknown, as we do not know the exact posterior probabilities $p(y|x)$. Replacing these by their estimates gives

$$\widehat{R}[x, \hat{f}(\hat{p}(1|x))] = \sum_{y \pm 1} L(y, \hat{f}(\hat{p}(1|x))) \, \hat{p}(y|x). \qquad (12)$$

Note that this formula does not take into account the inherent uncertainty in the estimate $\hat{p}(y|x)$. For example, a second-order distribution $G_x(q)$ with some spread and expectation of 1/2 implies that $p(1|x)$ *could* be much different from 1/2! In such cases, Eq. (12) is hence overly pessimistic.

A second-order distribution mitigates the over-pessimism in such regions. A more balanced estimate of the risk that takes into account a second-order distribution is

$$\widehat{ER}[x] = \mathbb{E}_q[\widehat{R}[x, \hat{f}(q)]] = \int_0^1 \widehat{R}[x, \hat{f}(q)] \, g_x(q) dq \qquad (13)$$

The intuition behind this estimate is as follows: if the second-order distribution has significant mass near both limits of its domain (i.e., it has a high density for values of $q = \hat{p}(y|x)$ close to 0 and 1), then it may be possible to construct a classifier with low risk at $x$, by querying additional labels in its neighborhood. As an extreme example, consider a second-order distribution for $\hat{p}(Y=1|x)$ given by Bernoulli(0.5), which implies that the conditional probability of class +1 is either 0 or 1. Then, $\widehat{R}[x] = 0.5$, but $\widehat{ER}[x] = 0$. This fact is taken into account by Eq. (13) but not by Eq. (12).

It is easy to prove that for any density $g_x(q)$, from which $\hat{p}(1|x)$ and $\hat{f}$ are derived via Eqs. (8) and (9), $\widehat{R}[x] \geq \widehat{ER}[x]$. Moreover, equality holds iff the entire mass of the second-order distribution lies on one side of the decision threshold, or if it is a Dirac distribution at 1/2. Interestingly, in these two cases it is of no benefit to query the label at $x$.

These properties suggest that the difference $\widehat{R}[x] - \widehat{ER}[x]$ is a good indicator for the potential importance of acquiring a label at $x$, though other choices seem possible. Consequently, taking also into account that the *overall* risk is a density-weighted mean of the local risk (see Eq. (10)), we propose the following training utility value (TUV):

$$TUV(x) = \left(\widehat{R}[x] - \widehat{ER}[x]\right) \cdot \hat{p}(x) \qquad (14)$$

where $\hat{p}(x)$ is some (non-parametric) estimate of the density at $x$. The weighting by the total density concentrates the learning effort on those regions of feature space that are actually relevant. Table 1 describes the pseudo-code of a single iteration of DEAL. Of course, the density estimate $\hat{p}(x)$ need be computed only once at the start of the AL process. Fig. 2 compares the TUV of DEAL to criteria used in experimental design and for US.

Table 1: One iteration of DEAL

| **Algorithm** DEAL |
|---|
| **Input:** Labeled set $\mathcal{L}$, unlabeled set $\mathcal{U}$. |
| **Output:** Selected sample $x \in \mathcal{U}$ and its label $y(x)$ |
| **Algorithm:** |
| 1: compute density estimate $\hat{p}(x)$ |
| 2: for all $x \in \mathcal{U}$ do |
|    - compute second-order distribution of $\hat{p}(Y=1\|x)$ by Eq. (6) |
|    - compute $TUV(x)$ by Eq. (14) |
| 3: query label $y$ of $x \in \mathcal{U}$ with largest $TUV$ |

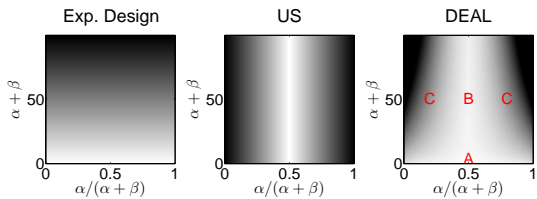

Figure 2: Training utility values, as a function of the two parameters $\alpha/(\alpha+\beta)$ and $\alpha+\beta$ in a second-order distribution of type $Beta(\alpha,\beta)$, for space-filling experimental design (Sacks et al., 1989), uncertainty sampling and DEAL. Roughly, $\alpha$ and $\beta$ measure the local amount of evidence for either class (Eqs. (6),(7)). Not taking sample density into account, the most interesting points for DEAL are those with little evidence for either class as yet (A), followed by points with evidence for both classes (B), followed by points with strong evidence for one and little for the other class (C). In contrast, US merely takes into account the distance from the decision boundary, pretending (A)≡(B). Space-filling experimental design prefers unexplored regions, regardless of their estimated class conditional probabilities, so that (B)≡(C).

## 5 RESULTS

We compare the empirical performance of DEAL to that of random sampling (RS), uncertainty sampling (US) (Lewis and Gale, 1994) and error reduction sampling (ERS) (Roy and McCallum, 2001). For a meaningful comparison, all methods use the same kernel density classifier, with an isotropic Gaussian kernel whose bandwidth is chosen according to the normal reference rule (Scott, 1992, chapter 6). The density $\hat{p}(x)$ in Eq. (14) is also estimated by kernel density estimation with the same kernel and bandwidth selection rule.

As is well known, non-parametric kernel density estimation with a limited number of samples may be highly inaccurate in high dimensions (Scott, 1992, chapter 7). Therefore, we first project the data to its $d$ leading principal components, where the dimension $d$ is chosen according to the resampling via permutation scheme of (Zhu and Ghodsi, 2006), with the minimum number of components set to two.

We always start with an empty set $\mathcal{L}$ of labeled points. In case of RS, US and ERS, the first query points are selected randomly until there is at least one label for each class. In case of DEAL, its deterministic strategy can be applied from the very beginning, with the first label requested for the point with the highest density estimate. In all our experiments, we set $\delta = 0.5$, consistent with Jeffreys' prior for the Bernoulli distribution (Jeffreys, 1946).

### 5.1 UCI DATA SETS BENCHMARK

We considered 32 of the most frequently used UCI data sets[3]. Each dataset was preprocessed as follows: $(i)$ Categorical variables with more than two outcomes were replaced by #outcomes$-1$ indicator variables, $(ii)$ missing values in categorical variables were treated as a separate outcome, $(iii)$ missing values in continuous inputs were replaced by the respective mean, and $(iv)$ the data was normalized to unit variance in each dimension. If a dataset had more than two classes, the classes were joined to create binary problems such that the new classes were approximately equally abundant.

---
[3]We excluded datasets with only categorical variables, or with significant missing data. We did not exclude datasets on which our AL scheme did not perform well.

All results are obtained from 10-fold cross validation (CV), i.e., nine tenths of the data were used in active learning, with one tenth reserved for testing. To average out the randomness of the initial labeling for RS, US and ERS, all experiments are repeated 5 times for each of the 10 CV folds. Fig. 3(a) shows

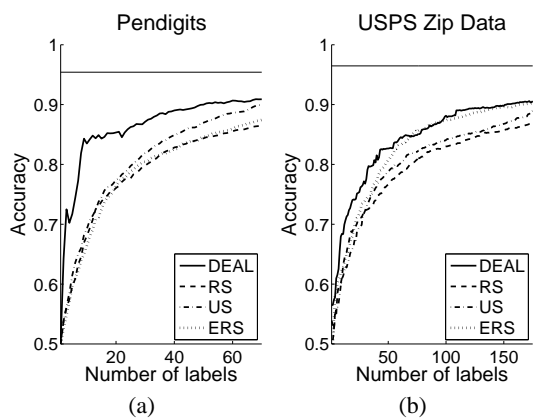

Figure 3: Learning curves for two data sets ("Pendigits" and "USPS, Grouping 10") where DEAL works well. Note that DEAL does not outperform the competing methods on every single data set, but on average across multiple data sets (see tables). The top horizontal line is the asymptotic performance of the classifier, at the end of the complete learning curve (estimated by 10-fold CV for UCI and on separate test set for the USPS data).

the learning curve for the Pendigits dataset. All others appear in the supplementary material.

To reward both initial steepness of the learning curve and early convergence to a high accuracy, we propose to measure performance by the area under the learning curve. All curves are truncated when the worst method achieves 90% of the accuracy of the classifier trained with the completely labeled training data[4], but at the latest after 200 iterations. As we compare only the relative performance

[4]Defined as the average of the 10-fold CV accuracies, each with a different set of 9/10 of the data fully labeled.

Table 2: Average accuracy of the compared AL strategies for 32 data sets from the UCI database with preprocessing as described in text, where $n$ is the total size of the data set and $d$ the dimension of the PCA subspace. The mean rank is computed based on ordering the performance of the AL strategies for each data set. The best and second best methods are indicated by bold font and italics, respectively. The mean rank test statistic is used for the statistical hypothesis tests described in the text.

| Dataset ($n$,$d$) | RS | US | ERS | DEAL |
|---|---|---|---|---|
| Anneal (898,17) | .813 | *.849* | .802 | **.857** |
| Audiology (226,9) | .664 | .650 | **.680** | *.666* |
| Autos (205,14) | *.653* | .638 | .614 | **.678** |
| Balance S. (625,2) | **.717** | .705 | *.716* | .715 |
| Breast C. (286,16) | *.644* | **.656** | .640 | .617 |
| Breast W (699,2) | .807 | *.835* | .820 | **.855** |
| Dermatol. (366,4) | .802 | *.841* | .789 | **.878** |
| Diabetes (768,2) | .684 | .682 | .687 | **.695** |
| Ecoli (336,3) | *.796* | .793 | .793 | **.852** |
| Glass (214,4) | .646 | **.688** | *.669* | .668 |
| Heart C (303,8) | .733 | .722 | *.748* | **.753** |
| Hepatitis (155,7) | .782 | *.796* | .781 | **.801** |
| Hyperth. (3772,11) | .863 | *.889* | .865 | **.919** |
| Ionosphere (351,5) | .782 | .802 | *.817* | **.841** |
| Iris (150,2) | .793 | *.809* | .807 | **.924** |
| Led 24 (1000,2) | .667 | .643 | .667 | **.695** |
| Letter (20000,5) | .631 | .627 | *.632* | **.652** |
| Liver (345,2) | .530 | **.541** | .516 | *.539* |
| Lymph (148,9) | .671 | **.712** | .681 | *.692* |
| Optdig. (5620,18) | .819 | *.849* | .791 | **.887** |
| Pendigits (7494,5) | .783 | *.804* | .783 | **.861** |
| Primary Tu (339,9) | *.652* | .650 | .647 | **.696** |
| Satimage (6435,3) | .777 | *.819* | .790 | **.852** |
| Segment (2310,3) | *.830* | .741 | .737 | **.871** |
| Sonar (208,8) | .695 | *.714* | .699 | **.725** |
| Soybean (683,20) | .786 | *.811* | .764 | **.831** |
| Vehicle (846,4) | .720 | **.736** | .721 | *.734* |
| Vote (435,8) | .803 | .799 | *.812* | **.841** |
| Vowel (990,16) | *.671* | .540 | .627 | **.694** |
| Waveform (5000,2) | .767 | **.793** | *.787* | *.787* |
| Wine (178,3) | .831 | .847 | *.856* | **.895** |
| Yeast (1484,2) | .571 | .562 | *.577* | **.592** |
| Mean Rank | 3.09 | 2.56 | 2.97 | 1.38 |

of different AL strategies for the same classification algorithm, this measure is equivalent to the one proposed in (Baram et al., 2004) and also used in (Schein and Ungar, 2007). The results for all data

Table 3: Average accuracy of the compared AL strategies for 10 different groupings of the USPS Zip Data with preprocessing as described in text. The best and second best method are indicated using bold font and italics, respectively. The mean rank is computed based on ordering the performance of the AL strategies for each grouping.

| Grouping | RS | US | ERS | DEAL |
|---|---|---|---|---|
| $\{1,2,3,4,5\}$ | 0.777 | 0.807 | *0.829* | **0.832** |
| $\{0,1,2,3,4\}$ | 0.786 | 0.808 | *0.831* | **0.837** |
| $\{1,3,5,7,9\}$ | 0.782 | 0.819 | **0.832** | *0.830* |
| $\{0,1,7,8,9\}$ | 0.774 | 0.811 | *0.817* | **0.830** |
| $\{1,3,4,5,9\}$ | 0.793 | 0.810 | *0.828* | **0.838** |
| $\{1,2,3,7,8\}$ | 0.782 | 0.797 | *0.825* | **0.833** |
| $\{0,1,6,8,9\}$ | 0.777 | 0.813 | *0.824* | **0.846** |
| $\{0,5,6,7,9\}$ | 0.777 | 0.805 | *0.815* | **0.830** |
| $\{0,2,4,5,8\}$ | 0.750 | 0.805 | *0.815* | **0.821** |
| $\{3,4,5,6,9\}$ | 0.791 | 0.799 | *0.825* | **0.840** |
| Mean Rank | 4.000 | 3.000 | 1.900 | 1.100 |

Table 4: Average accuracy of the compared AL strategies for 3 different groupings of the Caltech-4 data set with preprocessing as described in text. The best and second best method are indicated using bold font and italics, respectively.

| Grouping | RS | US | ERS | DEAL |
|---|---|---|---|---|
| $\{1,2\}$ vs. $\{3,4\}$ | 0.818 | *0.846* | 0.807 | **0.877** |
| $\{1,3\}$ vs. $\{2,4\}$ | 0.799 | *0.829* | 0.803 | **0.840** |
| $\{1,4\}$ vs. $\{2,3\}$ | 0.803 | *0.836* | 0.797 | **0.872** |
| Mean Rank | 3.333 | 2.000 | 3.667 | 1.000 |

sets are presented in Table 2.

We compare the performance of the different strategies as recommended in (Demšar, 2006). The Friedman test, which uses the mean performance ranks of Table 2, yields a $p$-value of $p = 2.53 \times 10^{-9}$ for the null hypothesis of equal performance of all strategies. For comparing all classifiers to each other, we use the two-tailed Nemenyi test. At a $1\%$ significance level its threshold for differences in Mean Rank is $1.004$. This means that DEAL performs significantly better than each of the other strategies. The performances of the other methods do not differ significantly from each other, even at the $10\%$ level (corresponding to a threshold of $0.739$).

### 5.2 USPS ZIP DATA

To obtain challenging classification tasks with convoluted decision boundaries, the digit images from the USPS corpus (LeCun et al., 1990) were grouped into two classes in various ways, see Table 3. All images were projected to the $d = 39$ leading principal components, with 7291 samples eligible for active learning and an independent set of 2007 samples held out for testing purposes.

As Table 3 shows, DEAL performed best in 9 out of the 10 groupings. Fig. 3(b) shows one learning curve, all others are in the supplementary material.

### 5.3 CALTECH-4

Caltech-4 is a well established standard benchmark for object categorization (Fergus et al., 2003) and has also been used in AL (Kapoor et al., 2007). This dataset consists of 4 different image groups: airplanes (category 1; 800 images), rear views of cars (2; 1155), frontal faces (3; 435) and motorbikes (4; 798). Fig. 4 shows one example from each category. We represent the images by the "Color and Edge Directivity Descriptor" (CEDD) (Chatzichristofis and Boutalis, 2008). The resulting 144-dimensional features were then projected to the 17 leading principal components. To create challenging two-class problems with convoluted decision boundaries, we grouped the 4 categories in three possible ways.

Table 4 presents the resulting performances, based on 10-fold CV with 5 repetitions (see Section 5.1). It shows that DEAL performs best for all groupings. Moreover, for this dataset, US is the second best strategy, probably because the problem is not as challenging as the Zip Data, that originally consists of 10 categories. Interestingly, ERS performs worse than random sampling in two out of three tasks.

## 6 DISCUSSION

In this paper we derived a new AL strategy, which considers not only the density and distance of an unlabeled sample to the decision boundary, but also the number of labeled points in its neighborhood. All this information is taken into account by requiring that the underlying classifier provide a distributional estimate at each unlabeled point, leading to a natural definition of its training utility value.

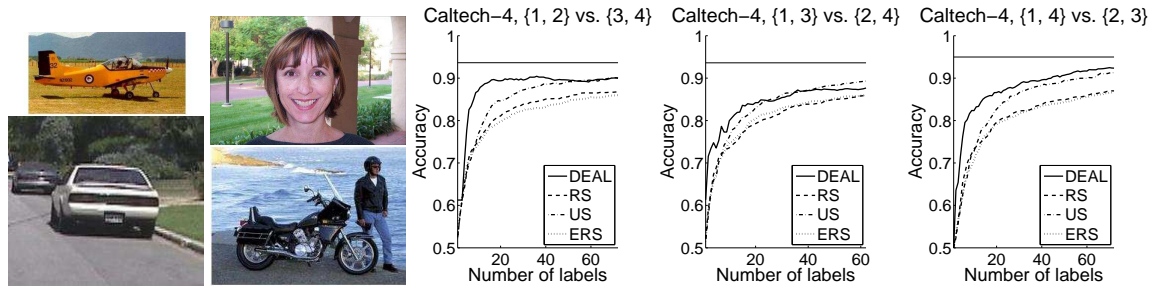

Figure 4: Left: Example images of the 4 object categories of Caltech-4 (airplane, car, face, motorbike). Right: Learning curves for three possible groupings of the 4 categories. DEAL performed best in all cases and US second best. Interestingly, ERS is not better than RS here (see also Table 4). The top horizontal line is the asymptotic accuracy of the classifier, with all training data labeled (estimated by 10-fold CV).

Information similar to that contained in a second-order distribution is *implicitly* used by methods that minimize the expected estimated risk (MacKay, 1992; Roy and McCallum, 2001; Zhu et al., 2003). These AL schemes indirectly measure the uncertainty of a point estimate, by perturbing the current classifier with hypothetical new labels and investing where the potential reduction in estimated risk is greatest.

In contrast, DEAL makes this dependence on the uncertainty explicit. Not only is it simple to implement, it also empirically outperformed error reduction sampling, uncertainty and random sampling schemes on a large collection of UCI, USPS and Caltech data sets. Note that one cannot expect a single strategy to perform best on all data sets. For instance, if the decision boundary is simple, strongly favoring exploitation over exploration (as in uncertainty sampling) may be the best strategy. For more challenging classification problems with complex class boundaries, balancing exploration and refinement, as DEAL does, seems a crucial ingredient for active learning.

While our AL scheme is general and applicable to any classifier that outputs second-order distributions, in this paper we focused for simplicity on its implementation with kernel density classification. As we shall describe in a future publication, second-order distributions can be derived for other classifiers, most notably random forest (Breiman, 2001). Encouragingly, with random forest as the base classifier, not only are lower classification errors achieved, but also the advantage of DEAL over the other AL strategies continues to hold.

Finally, we note that second-order distributions are not limited to AL. In the presence of few training data, they may be used to extend the "doubt" class to also include poorly explored regions with high uncertainty. While beyond the scope of this paper, second-order distributions are also useful for outlier detection, in applications such as optical inspection, where not all defects are known in advance when training the classifier. These extensions, as well as generalizing our AL scheme to multi-class problems, and deriving second-order distributions for other (discriminative) classifiers, are interesting topics for future research.

**Acknowledgements.**

BN was supported by a grant from the Israeli Science Foundation (ISF).